\documentclass[10pt,twocolumn,letterpaper]{article}

\usepackage{cvpr}              
\definecolor{cvprblue}{rgb}{0.21,0.49,0.74}
\usepackage[pagebackref,breaklinks,colorlinks,allcolors=cvprblue]{hyperref}

\def\confName{CVPR}
\def\confYear{2027}

\usepackage{balance}
\usepackage{booktabs}
\usepackage{multirow}
\usepackage{graphicx}
\usepackage{wrapfig}
\usepackage{bbm}
\usepackage{algorithm}
\usepackage{algpseudocode}

\title{Defending Vision-Language-Action Models against Transferable Attacks via Visual-Proprioceptive Consistency}

\author{Jiahao Wang\\
University of Arizona\\
{\tt\small jiahaow@arizona.edu}
\and
Ming F. Li\\
University of Arizona\\
{\tt\small lim@arizona.edu}
\and
Han Xu\\
University of Arizona\\
{\tt\small xuhan2@arizona.edu}
}

\begin{document}
\maketitle
\begingroup
\renewcommand{\thefootnote}{}
\footnotetext{This manuscript is a preprint.}
\endgroup
\begin{abstract}

Vision-Language-Action (VLA) models have demonstrated strong capabilities in a wide range of robotic tasks. However, adversarial attacks in VLA models have shown strong transferability across robotic tasks, posing significant threats to robotic systems. Despite their different attack strategies,  we observe that these transferable attacks share a common consequence: they disrupt the VLA model's perception towards the robot itself. Motivated by this observation, we propose a proprioception-grounded defense framework that leverages trusted proprioceptive information to detect and mitigate transferable adversarial attacks. Specifically, we utilize the robot's proprioceptive states to establish a trusted spatial reference, and compare it with the robot's perceived location inferred from visual features. Discrepancies between these two sources provide a reliable signal for detecting adversarial attacks. Extensive experiments demonstrate effective detection of transferable adversarial patches and improved closed-loop performance. Real-robot experiments further demonstrate the effectiveness of our defense on real-world robotic tasks.

\end{abstract}    
\section{Introduction}
\label{sec:intro}

\begin{figure*}[htbp]
    \centering
    \includegraphics[width=0.9\linewidth]{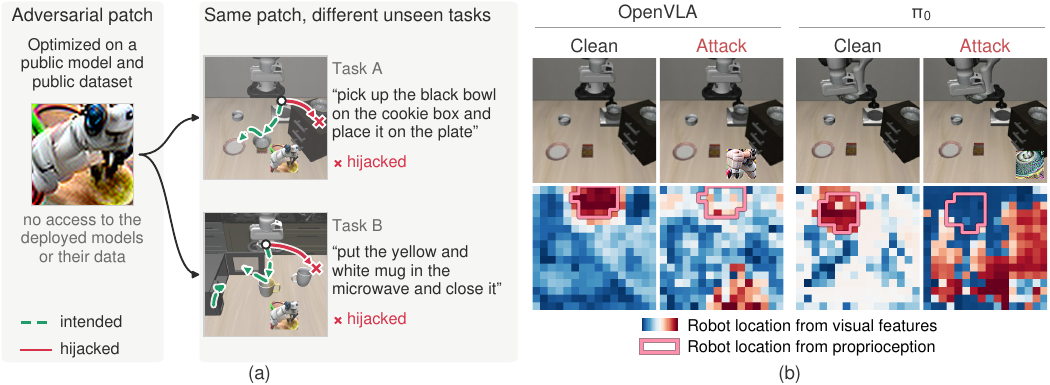}
    \caption{Transferable adversarial attacks against VLA models. \textbf{(a)} A single adversarial patch optimized using a public VLA model and source-task data can compromise fine-tuned VLA models on unseen downstream tasks. \textbf{(b)} Adversarial patches distort robot-related visual representations, causing the visually inferred robot location to deviate from its proprioception-based location.}
    \label{fig:motivation}
\end{figure*}

Vision-Language-Action (VLA) models~\cite{pmlr-v270-kim25c,black2024pi_0,intelligence2025pi_} leverage pretrained vision-language models to learn general-purpose robot policies, and have demonstrated strong capabilities in a wide range of robotic tasks. These models rely on camera observations to perceive their surroundings and extract visual features to guide action generation. However, prior studies have shown that vision models are usually vulnerable to adversarial perturbations~\cite{goodfellow2014explaining,brown2017adversarial,zhang2025adversarial}. Such vulnerabilities pose significant safety risks to VLA models, as adversarial perturbations can corrupt visual representations and consequently lead to incorrect robot actions.

More dangerously, recent studies have demonstrated that many adversarial attacks against VLA models can transfer across different robotic tasks~\cite{wang2025exploring,xu2025model,lu2026robots,fu2026vla}. As illustrated in Figure~\ref{fig:motivation}~(a), an adversarial patch can simultaneously cause failures in different tasks, even when these tasks are unknown to the attacker. In practice, attackers can develop such transferable patches using publicly available pretrained VLA checkpoints and datasets. If a VLA model is fine-tuned from one of these checkpoints, it is also likely to be vulnerable to the same attack, even if the attacker has no access to its fine-tuned parameters or private training data. Thus, this transferability could pose significant safety threats to real-world VLA deployment, as attackers can target a wide range of downstream tasks without developing task-specific attacks.

To better understand why adversarial attacks can transfer across different robotic tasks, we first conduct a visualization analysis to examine the VLA model's visual representations under attacks. Specifically, we analyze representative transferable attacks in Section~\ref{sec:preliminary} and identify two common mechanisms. \textbf{(1)} Some attacks broadly disrupt visual features, including the object and robot-related features to mislead the model's perception across different scenes and tasks~\cite{fu2026vla}. This mechanism is similar to universal adversarial attacks observed in general vision models~\cite{zhang2020understanding,zhang2021data,kim2024doubly}. \textbf{(2)} Other attacks target the perception towards the robot itself~\cite{wang2025exploring,xu2025model,lu2026robots}. Through explicitly or implicitly mimicking visual features associated with the robot, these attacks mislead the model's perception of the robot. Since accurate perception of the robot is important across different tasks, such attacks can also achieve cross-task transferability.

Despite their different mechanisms, we found that both types of attacks share a common consequence: \textbf{they distort the visual representation of the robot itself.} This observation reveals an important opportunity for defense: While adversarial patches can mislead a VLA model's visual perception of the robot, they cannot alter the robot's actual physical state. In fact, robots continuously measure their own states through proprioceptive sensors, which provide information such as joint positions and orientations independently of camera observations.  We can leverage this proprioception information to verify whether a VLA's visual perception of  the robot is reliable.

Therefore, we propose a proprioception-grounded defense framework that detects adversarial attacks by checking the consistency between proprioceptive information and visual representations. First, we use the robot's proprioceptive state, together with its kinematic model and camera calibration, to determine where the robot should appear in the camera scene. We then analyze the VLA model's visual features to estimate where the model perceives the robot. As shown in Figure~\ref{fig:motivation}~(b), these two estimates are consistent under benign conditions but diverge when attacks distort the robot's visual representation. By measuring this discrepancy, our framework can detect adversarial attacks without requiring any additional training or expensive inference. Upon detection, our framework also leverages this discrepancy to locate and resist the adversarial attacks.

We conduct extensive experiments on OpenVLA~\cite{pmlr-v270-kim25c} and $\pi_0$~\cite{black2024pi_0} across LIBERO~\cite{liu2023libero} task suites and various adversarial attacks. Results demonstrate that our framework can accurately detect various transferable attacks, outperforming existing detection baselines. Beyond detection, our framework effectively mitigates adversarial patches and significantly improves task success under attack. To further examine the robustness of our detection, we conduct two \textit{adaptive attacks} by preserving robot-related visual features and minimizing the visual-proprio inconsistency score. We find that both strategies reduce attack effectiveness, supporting the reliability of the defense design. \textit{Real-robot experiments} further demonstrate the effectiveness of our defense on real-world robotic tasks.

\noindent Our contributions are summarized as follows:
\begin{itemize}
    \item We identify a shared mechanism of transferable adversarial attacks in VLA: these attacks compromise robot-related visual representations. This observation motivates leveraging trusted robot proprioception to verify the reliability of visual perception.
\item We develop a principled, lightweight, and attack-agnostic defense framework that leverages visual-proprioceptive consistency checking to detect, localize, and mitigate adversarial attacks. Our approach requires no adversarial examples or additional model training and can be integrated into existing VLA models without modifying their parameters.  \textbf{To the best of our knowledge, it is the first work to leverage robot proprioceptive information for adversarial defense in VLA models.}
    \item Extensive experiments across two VLA models and a broad range of attacks validate the effectiveness and generalizability of our defense. Evaluations under benign visual variations and adaptive attacks, together with real-robot experiments, further validate its reliability.
\end{itemize}

\section{Preliminary}

In this section, we first introduce the basic notations and the threat model
considered in this work. Then, we review representative adversarial attacks in VLA models and conduct initial analysis on their attack mechanisms.

\subsection{Definitions and notations}

Vision-Language-Action (VLA) models learn policies that map visual observations and language instructions to executable robot actions. We define a given VLA model  as  $\pi_\theta$, which takes inputs $o_t = \left[\textbf{I}_t^{1},\cdots, \textbf{I}_t^{n}, l_t, \mathbf{q}_t \right]$ at time step $t$, where $\mathbf{I}_t^{i}$ is the image of the $i$-th camera view, $l_t$ is the language instruction, and $\mathbf{q}_t$ is the proprioceptive state.
The model predicts an action or action chunk: $\mathbf{A}_t = \left[\mathbf{a}_{t},\cdots,\mathbf{a}_{t+h-1}\right]$ to control the robot's motion. VLA models are typically trained on robot demonstrations collected across different tasks. VLA models employ Visual Transformer (ViT)-based visual encoders such as DINOv2~\cite{oquab2023dinov2} and SigLIP~\cite{zhai2023sigmoid} to extract visual representations from the input image. We denote the extracted visual features as $\mathbf{z}_t \in \mathbb{R}^{K \times d}$, where $K$ denotes the number of visual tokens and $d$ is the feature dimension. The feature of the $k$-th token is denoted by $\mathbf{z}_{t,k}$.

\subsection{Threat model and attack mechanism analysis}

\label{sec:preliminary}

\begin{figure}[t]
    \centering
    \includegraphics[width=\linewidth]{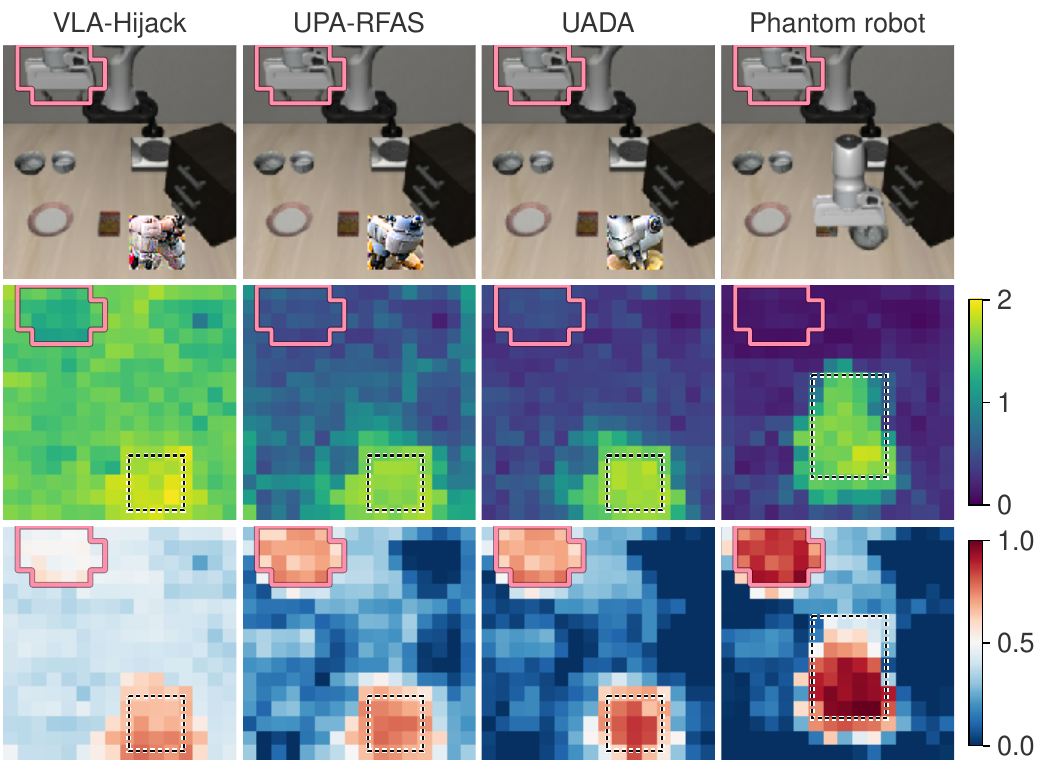}
    \caption{Visualization of representative attacks. 
    }
    \label{fig:visualization}
\end{figure}

\noindent \textbf{Threat Model.}  We focus on adversarial patch attacks~\cite{wang2025exploring,xu2025model,lu2026robots,fu2026vla} that can transfer across robotic tasks. As illustrated in Figure~\ref{fig:motivation} (a), a single adversarial patch can compromise multiple robotic tasks, even when these tasks are unknown to the attacker. Such attacks do not require access to target-task data or the victim model's fine-tuned parameters. Instead, the attacker optimizes an adversarial patch using a publicly accessible VLA model and source-task data $\mathcal{D}$ collected from diverse robotic tasks. The same patch can then be directly transferred to victim models fine-tuned from the public VLA checkpoint on unseen downstream tasks. Such transferability poses a significant threat to real-world VLA deployment, as a single adversarial patch can compromise a wide range of robotic applications.

There are several strategies to generate these patches. As a representative example, the work~\cite{wang2025exploring} introduces the Untargeted Action Discrepancy Attack (UADA), which aims to mislead the VLA model into generating incorrect robot actions. Specifically, UADA optimizes an adversarial patch to maximize the deviation between the predicted action $\mathbf{a}^{\mathrm{adv}}$ and the correct action $\mathbf{a}$:
\begin{equation}\label{eq:uada}
    \mathcal{L}_{\mathrm{UADA}} = \mathbb{E}_{\mathcal{D}}
    \Vert \mathbf{a}-\mathbf{a}^{\mathrm{adv}} \Vert_2^2,
\end{equation}
where $\mathcal{D}$ denotes the attacker's source data collected from diverse robotic tasks.

To better understand this attack, we extract visual features from the VLA model's vision encoder before and after applying the adversarial patch. We then measure the change in each visual token's representation $\mathbf{z}_k$.\footnote{The representation change for each visual token $k$ is measured as $1-\cos(\mathbf{z}_k,\mathbf{z}^{\mathrm{adv}}_k)$, where $\mathbf{z}_k$ and $\mathbf{z}^{\mathrm{adv}}_k$ denote its features under clean and adversarial observations, respectively.} As shown in Figure~\ref{fig:visualization} (middle), these changes are primarily concentrated around the patch location, while other regions remain largely unchanged. In Figure~\ref{fig:visualization} (bottom), we further examine the robot-likeness of visual features by measuring their similarity to robot-related features (see Section~\ref{sec:visual_consistency}). We find that the adversarial patch introduces strong robot-like features at its location. This suggests that the attack can mislead the VLA model's perception of the robot itself, even though its optimization objective directly targets the predicted actions.
Interestingly, this effect resembles introducing an additional robot into the observation to mislead the VLA model, which we refer to as a ``phantom robot'' in Figure~\ref{fig:visualization}. Some other attacks, such as UPA-RFAS~\cite{lu2026robots}, also exhibit similar effects.

In contrast, VLA-Hijack~\cite{fu2026vla} aims to suppress the VLA model's visual perception of the original robot. Specifically, it adopts a weighted visual feature disruption objective:
\begin{equation}\label{eq:supp}
    \mathcal{L}_{\mathrm{supp}} =
    \frac{\sum W_{\mathrm{robot}} \odot \mathrm{sim}(\mathbf{z},\mathbf{z}^{\mathrm{adv}})}{\sum W_{\mathrm{robot}}},
\end{equation}
where $W_{\mathrm{robot}}$ is a weighting matrix that assigns higher weights to robot-related regions. By minimizing this objective, the attack encourages greater feature disruption in these regions. Figure~\ref{fig:visualization} (bottom) shows that the robot-related features at the original robot location are substantially suppressed. Meanwhile, Figure~\ref{fig:visualization} (middle) reveals widespread changes in visual representations across the image. One possible explanation is that precisely suppressing robot-related features can be challenging, leading to broader feature disruption beyond the original robot region.

\section{Methodology}
\label{sec:method}

Analysis in Section~\ref{sec:preliminary} reveals that transferable adversarial patches disrupt or spoof robot-related visual representations. This inspires us to introduce \textbf{Visual-Proprio Consistency (VPC) Detection} to identify adversarial attacks by measuring the discrepancy between the robot's visual perception and its proprioceptive state. Upon detection, we further propose a gradient-based localization and mitigation strategy to identify and suppress suspicious visual regions.

\subsection{Proprioception-Guided Mask Generation}
\label{sec:mask}
\begin{figure}[h]
\centering
    \includegraphics[width=0.3\linewidth]{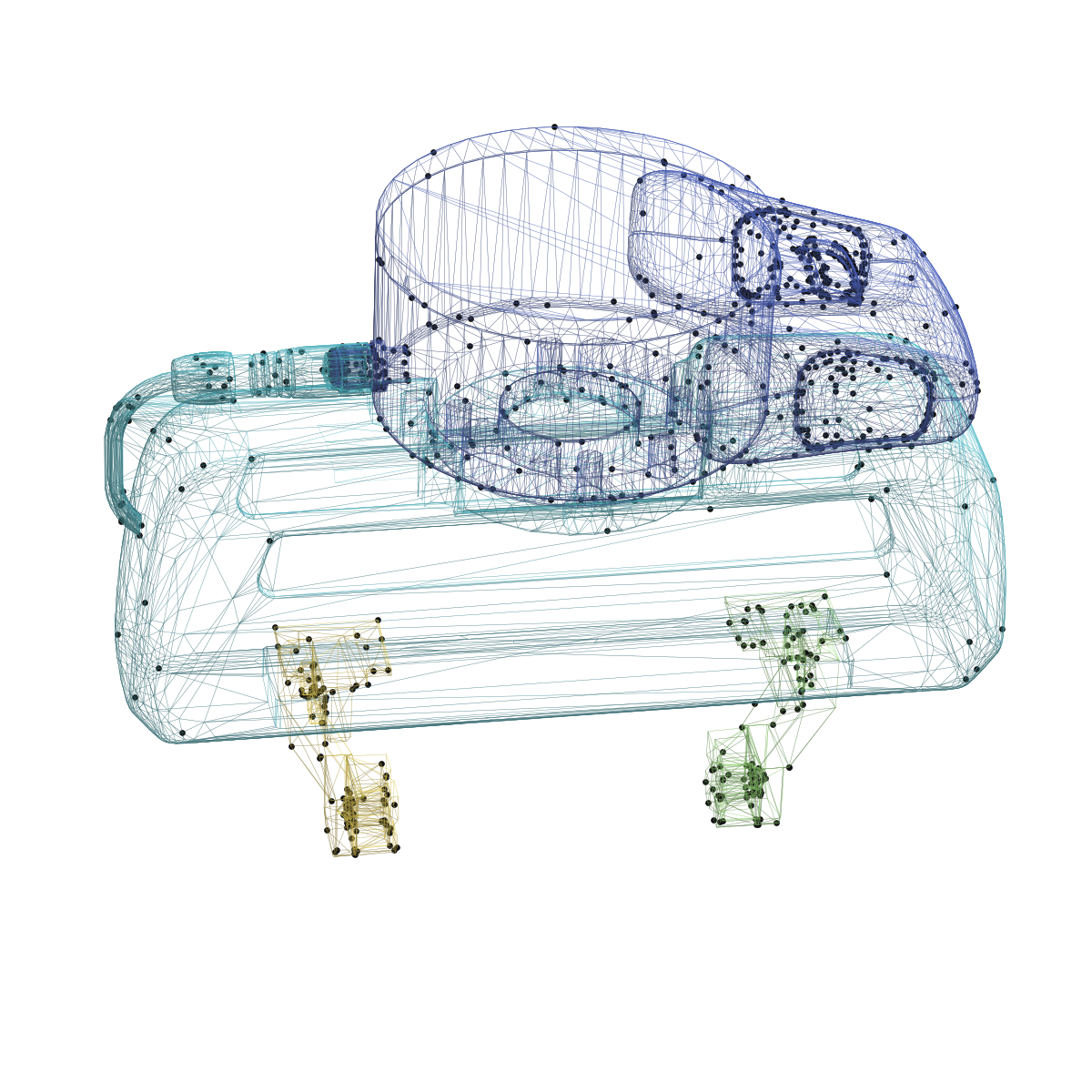}
    \includegraphics[width=0.3\linewidth]{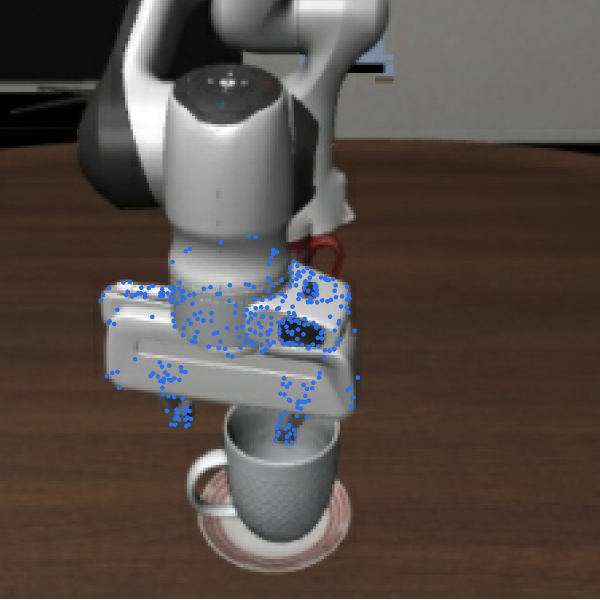}
    \includegraphics[width=0.3\linewidth]{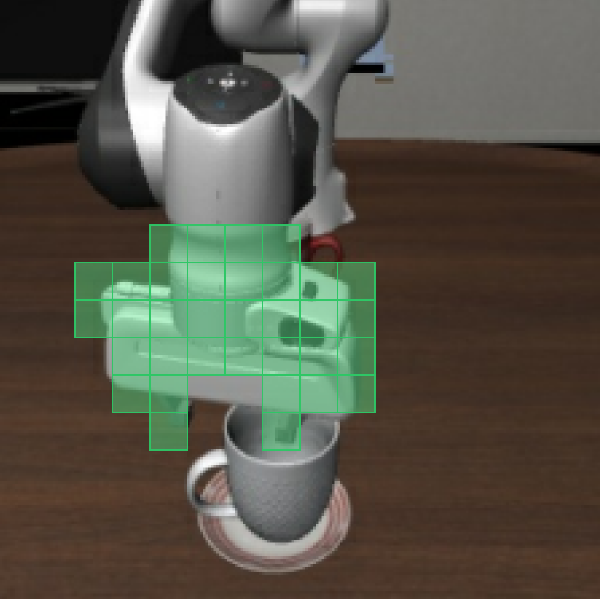}
    \caption{\textbf{Left:} Uniformly sample 3D points from the surfaces of the robot's components using their shapes and dimensions. \textbf{Middle:} 
    Project these points onto the camera image using calibrated camera parameters and proprioceptive state. \textbf{Right:} Generate a binary robot mask from the projected points.}
    \label{fig:mask_generation}
\end{figure}

To establish a trusted reference for the robot, we use its proprioceptive state to determine where it should appear in the camera image. As illustrated in Figure~\ref{fig:mask_generation}, we first obtain the 3D shapes and dimensions of the robot's components, which are typically provided by the robot manufacturer, and uniformly sample points on their surfaces. At timestep $t$, the robot's proprioceptive state $\mathbf{q}_t$ provides its current joint states, allowing us to calculate the 3D locations of these sampled points using forward kinematics. Since the camera's position, orientation, and imaging parameters are known from prior calibration, we can project these 3D points onto the camera image. 
Finally, we use the projected points to construct a robot mask $\mathbf{m}_t$, 
where $m_{t,k}$ indicates whether the $k$-th token belongs to the expected robot region:
\begin{align}
\label{eq:mask}
m_{t,k} =
\begin{cases}
1, & \text{if token } k \text{ belongs to the robot region},\\
0, & \text{otherwise}.
\end{cases}
\end{align}
Here, $k=1,\ldots,K$, where $K$ denotes the number of visual tokens. The resulting mask provides an independent spatial reference derived from the proprioception. More details could be found in Appendix~\ref{sec:appendix_mask}.

\subsection{Visual-Proprioceptive Consistency Detection}
\label{sec:visual_consistency}
We next aim to determine \textbf{where the VLA model perceives the robot in the image}. To this end, we examine the visual features extracted by the VLA's vision encoder to determine which image tokens exhibit robot-related features

\textbf{Robot Feature Distribution.} To determine whether a visual token represents the robot, a straightforward way is to measure its similarity to a reference robot feature. However, the robot's visual features can vary across different configurations, camera viewpoints, and visual conditions. A single reference feature may not adequately represent the robot across different observations. Therefore, we model robot-related visual features using a Gaussian distribution.
Specifically, given a set of clean observations from the training data, we extract visual features using the VLA model's vision encoder. For each observation, we collect robot-related features from tokens with $\mathbf{m}_{t,k}=1$, according to Eq.~\ref{eq:mask}. We then compute their mean $\boldsymbol{\mu}_R$ and covariance $\boldsymbol{\Sigma}_R$, which characterize the typical robot features and their variations. We refer to the resulting Gaussian distribution $\mathcal{N}(\boldsymbol{\mu}_R,\boldsymbol{\Sigma}_R)$ as the \emph{robot feature distribution (R-distribution)}.

 During model inference, we need to locate the robot from visual features. To achieve our goal, we aim to determine whether a visual token is associated with the robot by measuring its compatibility with the R-distribution. Specifically, we use the squared Mahalanobis distance to measure how well each token's visual feature fits the R-distribution:
\begin{equation}    \mathcal{D}_R(\mathbf{z}_{t,k}) =
    (\mathbf{z}_{t,k} - \boldsymbol{\mu}_R)^\top
    \boldsymbol{\Sigma}_R^{-1}
    (\mathbf{z}_{t,k} - \boldsymbol{\mu}_R).
    \label{eq:robot_distance}
\end{equation}
A smaller distance indicates that the visual token is more consistent with the distribution of robot-related features. We therefore use this distance to assess how strongly each visual token matches the robot's visual representation.

\textbf{Distinguishing Robot from Background.}
However, the semantic information captured by visual features can differ substantially across vision encoders and feature layers. For example, features from deeper layers may become increasingly similar across various locations, making it difficult to distinguish robot tokens using Eq.~\ref{eq:robot_distance} alone. To reduce the dependence on specific encoders and layers, we introduce a background feature distribution (B-distribution) as an additional reference. By contrasting each token's distance to the two distributions, we adopt a discriminative matching strategy that evaluates its relative compatibility with robot and background features. We construct the \emph{background feature distribution (B-distribution)} $\mathcal{N}(\boldsymbol{\mu}_B,\boldsymbol{\Sigma}_B)$ using the same procedure as the R-distribution, but collecting visual features from tokens with $\mathbf{m}_t=0$. We then compute the squared Mahalanobis distance $\mathcal{D}_B(\mathbf{z}_{t,k})$ to the B-distribution and define the robot-likeness score as the difference between the two distances:
\begin{equation}
    \mathcal{R}_{t,k} = \mathcal{D}_B(\mathbf{z}_{t,k}) - \mathcal{D}_R(\mathbf{z}_{t,k}).
\end{equation}
A higher score indicates that the token's visual feature is more consistent with the robot distribution than with the background distribution.

Finally, we apply spatial softmax to the scores $\mathcal{R}_{t,k}$ to obtain a normalized visual localization distribution $\boldsymbol{\pi}_t$, where $\boldsymbol{\pi}_{t,k}=\mathrm{softmax}_k(\mathcal{R}_{t,k}/T)$ and $T$ is the temperature parameter. This distribution indicates where the VLA's visual features suggest the robot is located (we use this as the robot-likeness scores in Figure~\ref{fig:visualization} (bottom)).

\textbf{Attack Detection.} 
If the robot is correctly perceived, visual tokens with high robot-likeness should be concentrated within the region predicted by proprioception. To quantify deviations from this expectation, we compute the distance from each visual token to the nearest token with $m_{t,k}=1$. We then aggregate these distances weighted by the visual localization distribution \(\boldsymbol{\pi}_t\):
\begin{equation}
    D_{\mathrm{PV}}(\mathbf{x}_t,\mathbf{q}_t)
=
\sqrt{
\sum_{k=1}^{K}
\pi_{t,k}
\min_{j:m_{t,j}=1}
\|\mathbf{p}_k-\mathbf{p}_j\|_2^2
},
\end{equation}
where $\mathbf{p}_k$ denotes the spatial center of visual token $k$. A larger discrepancy indicates that the robot-related visual evidence deviates further from its expected location. We therefore detect an attack using $D_{\mathrm{PV}}$ as an indicator of the visual-proprioceptive inconsistency.

\subsection{Consistency-Guided Patch Localization and Mitigation}

Once an adversarial attack is detected, our goal is to suppress the patch's influence while preserving task-relevant visual information. To achieve this, we first identify suspicious image regions and then selectively suppress their influence through attention masking. Specifically, we leverage the differentiability of $D_{\mathrm{PV}}$ and use gradient-based attribution to identify regions that contribute most to the inconsistency. To reduce the noise from the gradients and obtain a more accurate estimation, we adopt a SmoothGrad~\cite{smilkov2017smoothgrad} based method, which averages gradients over multiple noisy perturbations of the input. Given an input image $\mathbf{I}_t$, we compute the attribution map as
\begin{equation}
    A(u, v) = \left\Vert \frac{1}{L} \sum_{l=1}^L \frac{\partial D_{PV}(\mathbf{I}_t + \mathbf{\eta}^{(l)}, \mathbf{q}_t)}{\partial \mathbf{I}_t (u,v,:)} \right\Vert_1,
\end{equation}
where $\mathbf{\eta}^{(l)} \sim \mathcal{N}(\mathbf{0}, \sigma^2\mathbf{I})$, \(L\) is the number of noisy samples, and \((u,v)\) denotes the pixel coordinates. Notably, high attribution values appear in both the adversarial patch and the actual robot region, as both can strongly influence the consistency score. To distinguish the patch from the robot itself, we leverage the proprioception-guided mask to exclude the expected robot region from the attribution map. 

Based on the attribution map obtained during localization, we rank the visual tokens according to their contribution to the detected inconsistency. We progressively mask the highest-ranked tokens and recompute $D_{\mathrm{PV}}$ after each step. If the inconsistency remains above the detection threshold $\delta$, we continue expanding the masked token set until the score falls below $\delta$ or a predefined masking budget is reached. This progressive strategy limits the extent of visual information suppression while reducing the detected inconsistency. To implement token suppression without modifying the input image, we mask the corresponding Key positions by adding a negative bias to the pre-softmax attention logits:
\begin{equation*}
\widetilde{\mathbf A} = \operatorname{Softmax}\left(\frac{\mathbf Q\mathbf K^\top}{\sqrt d}+\mathbf B \right), B_{ij}= \begin{cases} -\infty, & j\in\mathcal H_t,\\ 0, & \text{otherwise}, \end{cases}
\end{equation*}
where $\mathcal H_t$ denotes the currently masked tokens. This masking is applied across all attention layers of the visual encoders, preventing the selected tokens from contributing to attention aggregation.

\section{Experiments}

We evaluate our defense across different VLA architectures, manipulation tasks, and adversarial attacks. We first compare VPC with existing vision-only defenses to highlight the benefits of incorporating robot proprioception. We then assess its ability to restore task performance through closed-loop evaluations in simulation and the real world. Finally, we evaluate robustness under natural visual variations and adaptive attacks.

\paragraph{Models and benchmarks.}
We evaluate our method on two representative open-source VLA models, OpenVLA~\cite{pmlr-v270-kim25c} and $\pi_0$~\cite{black2024pi_0}, primarily in the LIBERO~\cite{liu2023libero} simulation environment across four task suites: Spatial, Object, Goal, and Long. We further validate our defense on a real-world robotic platform equipped with a WidowX AI robotic arm and an Intel RealSense D455f camera (in Section~\ref{sec:real}).

\vspace{0.2cm}
\noindent\textbf{Considered Attacks.}
We evaluate three adversarial patch attacks, VLA-Hijack~\cite{fu2026vla}, UPA-RFAS~\cite{lu2026robots}, and EDPA~\cite{xu2025model}, on both models. We additionally evaluate four patch variants: UADA, UMA, UPA, TMA from the work~\cite{wang2025exploring} on OpenVLA. These attacks are designed for OpenVLA's discretized action predictions and cannot be directly applied to $\pi_0$, which generates continuous action chunks through flow matching. In our study, we optimize these attacks only in LIBERO-\textbf{Spatial} task suite. As shown in Table~\ref{tab:attack}, they could also transfer to other task suites to cause task success rate drop.\footnote{We observe UPA-RFAS and EDPA have limited transferability on $\pi_0$. However, VLA-Hijack and Phantom remain effective, indicating that $\pi_0$ is still vulnerable to transferable adversarial attacks.} Beyond these adversarial patches, we introduce a phantom-robot attack that adds an additional robotic arm (the same as in Figure~\ref{fig:visualization}).

\begin{table}[htbp]
  \caption{Task success rate under various attacks on LIBERO for OpenVLA (top) and $\pi_0$ (bottom).}
  \vspace{-0.3cm}
  \label{tab:attack}
  \centering
  \begin{tabular}{@{}lcccc@{}}
    \toprule
    Attack & Spatial & Goal & Object & Long \\
    \midrule
    VLA-Hijack & 0.00 & 0.01 & 0.00 & 0.07\\
    UPA-RFAS & 0.00 & 0.00 & 0.00 & 0.00\\
    $\mathrm{UMA}$ &0.00 & 0.01& 0.00& 0.00\\
    $\mathrm{UADA}$ &0.00 & 0.00& 0.00& 0.00\\
    $\mathrm{UPA}$ &0.06 & 0.10& 0.02& 0.01\\
    $\mathrm{TMA}$ &0.00 & 0.00& 0.00& 0.00\\
    EDPA &0.00 & 0.00& 0.00& 0.01\\
    Phantom & 0.06 & 0.02 & 0.07 & 0.01 \\
    Clean & 0.78 & 0.68 & 0.73 & 0.53 \\
    \midrule
    VLA-Hijack & 0.12 & 0.40 & 0.25 & 0.08\\
    UPA-RFAS$^*$ & 0.53 & 0.85 & 0.97 & 0.70 \\
    EDPA$^*$ & 0.72 & 0.94 & 0.95 & 0.65 \\
    Phantom & 0.59 & 0.36 & 0.61 & 0.25 \\
    Clean & 0.95 & 0.93 & 0.95 & 0.85 \\
    \bottomrule
  \end{tabular}
\end{table}

\begin{table*}[t]
  \caption{Adversarial attacks detection comparison (AUROC) on OpenVLA (top) and $\pi_0$ (bottom).}
  \vspace{-0.3cm}
  \label{tab:comparison}
  \centering
  \resizebox{0.95\textwidth}{!}{
  \begin{tabular}{@{}l*{12}{c}@{}}
    \toprule
    \multirow{2}{*}{Attack}
      & \multicolumn{2}{c}{Deep kNN}
      & \multicolumn{2}{c}{Feature Squeezing}
      & \multicolumn{2}{c}{ARMOR}
      & \multicolumn{2}{c}{Jedi}
      & \multicolumn{2}{c}{LGS}
      & \multicolumn{2}{c}{Ours} \\
    \cmidrule(lr){2-3}
    \cmidrule(lr){4-5}
    \cmidrule(lr){6-7}
    \cmidrule(lr){8-9}
    \cmidrule(lr){10-11}
    \cmidrule(lr){12-13}
      & Avg. & Worst
      & Avg. & Worst
      & Avg. & Worst
      & Avg. & Worst
      & Avg. & Worst
      & Avg. & Worst \\
    \midrule

    VLA-Hijack
      & 0.990 & 0.961
      & 0.932 & 0.915
      & 0.954 & 0.833
      & 0.953 & 0.849
      & 0.931 & 0.813
      & \textbf{0.992} & \textbf{0.968} \\

    UPA-RFAS
      & 0.979 & 0.921
      & 0.923 & 0.913
      & 0.853 & 0.556
      & 0.929 & 0.841
      & 0.699 & 0.289
      & \textbf{0.999} & \textbf{0.995} \\

    UMA
      & 0.958 & 0.834
      & 0.910 & 0.895
      & 0.941 & 0.889
      & 0.984 & 0.944
      & 0.458 & 0.082
      & \textbf{0.999} & \textbf{0.997} \\

    UADA
      & 0.952 & 0.824
      & 0.921 & 0.911
      & 0.829 & 0.696
      & 0.951 & 0.830
      & 0.752 & 0.429
      & \textbf{0.998} & \textbf{0.995} \\

    UPA
      & 0.972 & 0.890
      & 0.903 & 0.868
      & 0.857 & 0.696
      & 0.929 & 0.841
      & 0.699 & 0.289
      & \textbf{0.990} & \textbf{0.976} \\

    TMA
      & 0.942 & 0.789
      & 0.922 & 0.910
      & 0.959 & 0.950
      & 0.983 & 0.941
      & 0.379 & 0.132
      & \textbf{0.998} & \textbf{0.995} \\

    EDPA
      & 0.994 & 0.978
      & 0.921 & 0.899
      & 0.851 & 0.635
      & 0.979 & 0.936
      & 0.445 & 0.131
      & \textbf{0.995} & \textbf{0.979} \\

    Phantom
        & 0.781 & 0.694
        & 0.533 & 0.520
        & 0.509 & 0.434
        & 0.501 & 0.465
        & 0.502 & 0.436
        & \textbf{0.974} & \textbf{0.911}\\

    \midrule

    VLA-Hijack
      & \textbf{1.000} & \textbf{1.000}
      & 0.995 & 0.992
      & 0.921 & 0.893
      & 0.931 & 0.754
      & \textbf{1.000} & \textbf{1.000}
      & 0.998 & 0.995 \\

    UPA-RFAS
      & \textbf{0.998} & \textbf{0.996}
      & 0.566 & 0.540
      & 0.439 & 0.365
      & 0.971 & 0.890
      & 0.765 & 0.412
      & 0.986 & 0.963 \\

    EDPA
      & \textbf{1.000} & \textbf{1.000}
      & 0.711 & 0.693
      & 0.513 & 0.427
      & 0.985 & 0.941
      & 0.765 & 0.415
      & 0.998 & 0.994 \\

    Phantom
      & 0.713 & 0.663
      & 0.621 & 0.564
      & 0.477 & 0.377
      & 0.497 & 0.408 
      & 0.543 & 0.417
      & \textbf{0.950} & \textbf{0.920}\\

    \bottomrule
  \end{tabular}
  }
\end{table*}

\subsection{Frame-Level Adversarial Detection}
\label{sec:frame_detection}

We first aim to answer the core question: Given a fixed scene, can our VPC defense effectively detect adversarial attacks, and how does it compare with existing defenses? To this end, we evaluate OpenVLA and $\pi_0$ across all four LIBERO suites under diverse attacks, comparing our method with existing detection baselines. We focus on frame-level detection here, leaving the evaluation of closed-loop defense performance to Section~\ref{sec:close}.

\vspace{0.2cm}
\noindent\textbf{Baseline Defenses.} We select five representative \textbf{vision-only defenses that rely solely on visual information,} allowing us to evaluate the benefits of incorporating robot proprioception. Deep kNN~\cite{sun2022out} detects out-of-distribution inputs using feature-space distances, while Feature Squeezing~\cite{xu2017feature} identifies adversarial examples through prediction inconsistencies under input transformations. ARMOR~\cite{pmlr-v202-liu23n} exploits anomalous attention patterns in ViTs to locate and mask adversarial patches. Jedi~\cite{tarchoun2023jedi} and LGS~\cite{naseer2019local} identify suspicious regions using image entropy and local gradients, respectively, and mitigate their influence through image restoration or smoothing.

\vspace{0.2cm}
\noindent\textbf{Results.} For each LIBERO task suite, we evaluate 280 clean observations and 280 adversarial observations for each attack (40 episodes in each suite and 7 frames in each episode). We use AUROC to assess the ability to distinguish clean from attacked observations. Table~\ref{tab:comparison} reports the mean and minimum AUROC across the four task suites for each attack, reflecting overall detection performance and performance in the most challenging suite, respectively. Our method achieves the highest mean and minimum AUROC on OpenVLA, while remaining highly effective on $\pi_0$.

\begin{figure}[h]
    \centering
\vspace{-0.3cm}
    \includegraphics[width=0.9\linewidth]{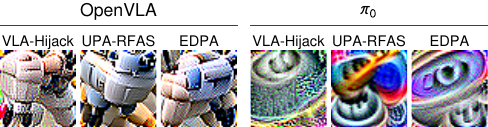}
    \vspace{-0.3cm}
    \caption{Visualization of adversarial patches.}
    \label{fig:patch}
\end{figure}

For the baselines, Deep kNN achieves high AUROC on $\pi_0$ under adversarial patch attacks, but its performance drops substantially on OpenVLA. As shown in Figure~\ref{fig:patch}, patches optimized for $\pi_0$ often exhibit visually irregular patterns, potentially producing distinctive out-of-distribution features that Deep kNN can readily detect. In contrast, patches targeting OpenVLA appear smoother and more visually natural, making them potentially harder to distinguish from benign features. This limitation becomes more evident under the phantom-robot attack, which introduces an additional robotic arm with a natural appearance into the observation. Similarly, Jedi, LGS, and ARMOR rely on specific visual cues, such as anomalous local entropy, gradients, or attention concentration, which may not consistently appear across different attacks. Unlike these vision-only defenses, VPC detects inconsistencies between visual representations and trusted robot proprioception, \textbf{making it less dependent on attack-specific visual characteristics.}

\vspace{-0.2cm}

\subsection{Closed-Loop Defense Performance}\label{sec:close}

We next evaluate whether detection-guided intervention improves closed-loop task execution under attack. Using the LIBERO benchmark, we evaluate 100 episodes per suite with adversarial attacks detected by our VPC detector. Table~\ref{tab:close_loop} shows the closed-loop task success rate under various attacks with and without our defense method to show the defense's effectiveness. We also report the performance for clean observations to validate that our method does not undermine the performance for benign observations. Results show that our method substantially improves task success rates under adversarial attacks, while preserving clean task performance. These results demonstrate that our attack-agnostic defense can effectively mitigate adversarial patches and restore closed-loop task performance across different VLA architectures and manipulation tasks. 

\begin{table}[h]
\caption{Closed-loop task success rate (\%) on OpenVLA (top) and $\pi_0$ (bottom), with and without defense.}
  \centering
  \resizebox{0.9\columnwidth}{!}{
  \begin{tabular}{@{}l*{8}{c}@{}}
    \toprule
    \multirow{2}{*}{Attack}
      & \multicolumn{2}{c}{Spatial}
      & \multicolumn{2}{c}{Object}
      & \multicolumn{2}{c}{Goal}
      & \multicolumn{2}{c}{Long} \\
    \cmidrule(lr){2-3}
    \cmidrule(lr){4-5}
    \cmidrule(lr){6-7}
    \cmidrule(lr){8-9}
      & w/ & w/o
      & w/ & w/o
      & w/ & w/o
      & w/ & w/o \\
    \midrule

    VLA-Hijack
      & 64 & 0
      & 73 & 0
      & 70 & 1
      & 42 & 7 \\

    UPA-RFAS
      & 67 & 0
      & 70 & 0
      & 69 & 0
      & 46 & 0 \\

    UMA
      & 58 & 0
      & 67 & 0
      & 66 & 1
      & 51 & 0 \\

    UADA
      & 69 & 0
      & 75 & 0
      & 68 & 0
      & 45 & 0 \\

    UPA
      & 67 & 6
      & 64 & 2
      & 68 & 10
      & 47 & 1 \\

    TMA
      & 68 & 0
      & 71 & 0
      & 71 & 0
      & 48 & 0 \\

    EDPA
      & 68 & 0
      & 67 & 0
      & 68 & 0
      & 45 & 1 \\
    
    Phantom
      & 67 & 6
      & 64 & 7
      & 70 & 2
      & 40 & 1 \\
      
    Clean
      & 79 & 78
      & 73 & 73
      & 65 & 68
      & 55 & 53 \\

    \toprule

    VLA-Hijack
      & 84 & 12
      & 97 & 25
      & 94 & 40
      & 82 & 8 \\

    UPA-RFAS
      & 92 & 53
      & 98 & 97
      & 92 & 85
      & 84 & 70 \\

    EDPA
      & 94 & 72
      & 98 & 95
      & 94 & 94
      & 83 & 65 \\

    Phantom
      & 83 & 59
      & 95 & 61
      & 81 & 36
      & 73 & 25 \\

    Clean
      & 94 & 95
      & 94 & 95
      & 94 & 93
      & 85 & 85 \\

    \bottomrule
  \end{tabular}
  }
  
  \label{tab:close_loop}
\end{table}

\subsection{Real-World Experiments}\label{sec:real}
\begin{figure}
    \centering
    \includegraphics[width=\linewidth]{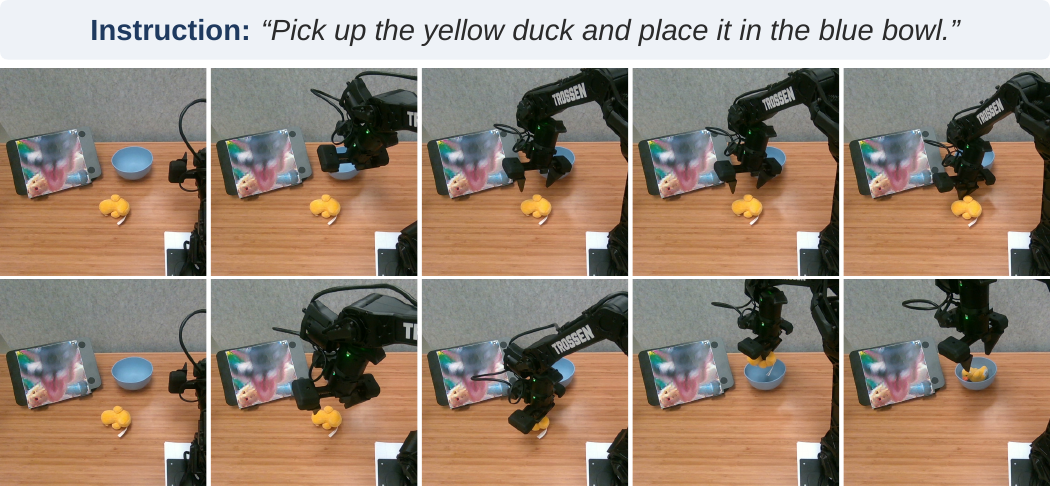}
    \caption{Real-World Evaluation. \textbf{Top:} The VLA model fails to complete the task without defense due to the patch. \textbf{Bottom:} The VLA model successfully completes the task with defense.}
    \label{fig:real_robot}
\end{figure}

To further validate the practical effectiveness of our defense, we deploy our framework with OpenVLA on a real-world robotic platform consisting of a WidowX AI robotic arm and an Intel RealSense D455f camera. The robot is instructed to ``pick up the yellow duck and place it in the blue bowl''. Under clean conditions, the robot succeeds in 19 out of 20 trials. We then introduce a representative UPA~\cite{wang2025exploring} attack, as illustrated in Figure~\ref{fig:real_robot}. Under this attack, the robot succeeds in only 5 out of 20 trials without defense. With our defense enabled, all 20 attacked trials are successfully detected, and our defense enables the robot to successfully complete 16 of them. These results demonstrate the effectiveness of our defense in mitigating adversarial patches and restoring task performance in real-world scenarios.

\subsection{Robustness to Natural Visual Variations}
\label{sec:robust}

\begin{figure}[htbp]
    \centering
    \vspace{-0.4cm}
    \includegraphics[width=0.9\linewidth]{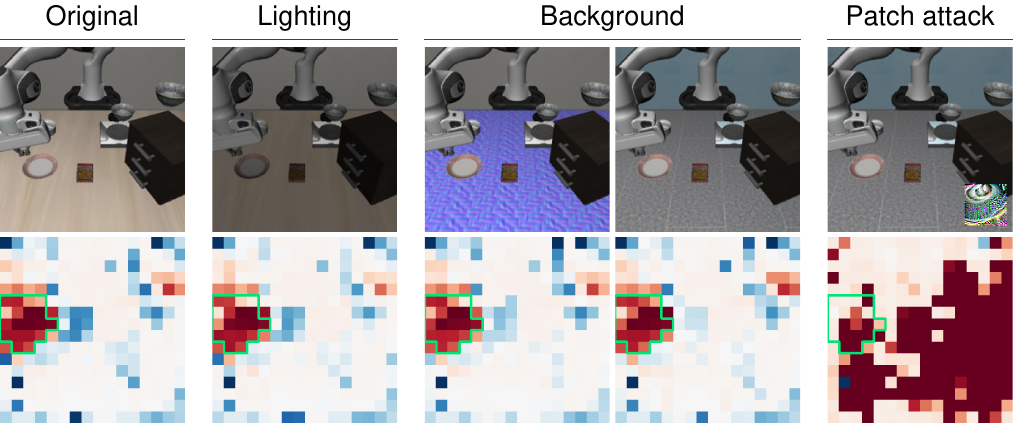}
    \caption{Detection under natural visual variants.}
    \label{fig:light_shift}
\end{figure}

In practical deployments, changes in lighting and background are common and may substantially alter visual representations. To evaluate whether our method remains effective under such changes, we conduct experiments using the lighting and background variations provided by LIBERO-Plus~\cite{Fei_2026_CVPR}. Under this setting, we retain the detector parameters learned from the original LIBERO dataset without retraining or adaptation to the shifted environments. Figure~\ref{fig:light_shift} visualizes the Mahalanobis-distance-based robot localization scores under natural visual variations and adversarial attacks. Despite the changes in visual appearance, our method continues to accurately localize the robot under benign conditions, while adversarial patches cause substantial localization deviations.  Results in Table~\ref{tab:light_shift} show that our method remains robust under benign visual variations, and can still reliably detect adversarial attacks.

\begin{table}[htbp]
  \caption{Detection performance (AUROC) under natural visual variation on OpenVLA (top) and $\pi_0$ (bottom).}
  \vspace{-0.3cm}
  \label{tab:light_shift}
  \centering
  \begin{tabular}{@{}lcccc@{}}
    \toprule
    Attack & Spatial & Goal & Object & Long \\
    \midrule
    VLA-Hijack & 1.000 & 0.999 & 0.998 & 0.905 \\
    UPA-RFAS & 0.996 & 0.999 & 0.999 & 0.992 \\
    UMA &1.000 & 0.998 &0.998& 0.993\\
    UADA &1.000 & 0.997& 0.999& 0.990\\
    UPA &1.000 & 0.985& 0.991&0.945\\
    TMA &1.000 & 0.998& 0.999& 0.990\\
    EDPA &1.000 & 1.000& 0.999& 0.976\\
    Phantom & 0.980 & 0.973 & 0.990 & 0.983  \\
    \midrule
    VLA-Hijack & 1.000 & 0.992 & 1.000 & 0.994\\
    UPA-RFAS & 0.985 & 0.977 & 1.000 & 0.962 \\
    EDPA & 0.997 & 0.995 & 1.000 & 0.996 \\
    Phantom & 0.979& 0.988& 0.942& 0.910 \\
    \bottomrule
  \end{tabular}
\end{table}

\subsection{Adaptive Attacks}

We further investigate whether an adaptive attacker can evade our defense while maintaining attack effectiveness. To this end, we consider two variants of UADA~\cite{wang2025exploring} with additional constraints on robot-related visual features and our detection criterion, respectively. We optimize and test these patches against OpenVLA on LIBERO-Spatial with different constraint weights $\lambda$.

\begin{figure}
    \centering
    \includegraphics[width=0.85\linewidth]{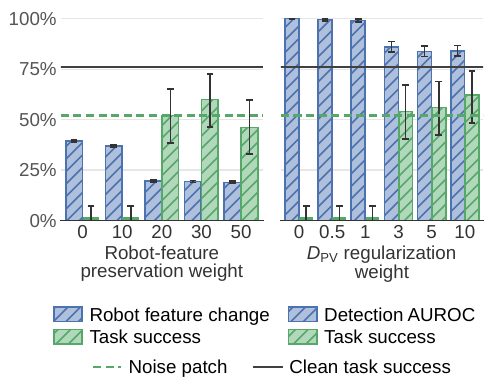}
    \vspace{-0.2cm}
    \caption{Adaptive attacks. \textbf{Left:}Robot feature preserving attack. \textbf{Right:} Detection-aware attack.}
    \label{fig:adaptive}
\end{figure}

\vspace{0.2cm}
\noindent\textbf{Feature-Preserving Attack.} We first examine whether an attacker can effectively manipulate a VLA policy while preserving robot features. Specifically, we measure the normalized squared $L_2$ distance between the visual features of robot tokens in clean and attacked images, averaged over all training samples. The attack performance is shown in Figure~\ref{fig:adaptive} (left), the dash lines show results under a random noise patch. When $\lambda$ is set small, the attack distorts more robot visual features and achieves a high attack success rate. While when we increase $\lambda$, though more robot visual features are preserved,  the attack success rate becomes much lower to be around a noise patch baseline. These results suggest that robot-feature disruption is an important contributor to attack effectiveness: constraining such disruption substantially weakens the attack. These findings support us to leverage the visual-proprio consistency to defend against these transferable attacks.

\vspace{0.2cm}
\noindent\textbf{Detection-Aware Attack.} We further consider a stronger adaptive setting in which the attacker has full knowledge of our detection mechanism. Specifically, we directly incorporate the proposed discrepancy score $D_{\mathrm{PV}}$ into the UADA optimization objective, encouraging the patch to evade detection. As shown in Figure~\ref{fig:adaptive} (right), we observe a similar trade-off as above: weak regularization preserves attack effectiveness but fails to evade detection, whereas stronger regularization reduces detectability at the cost of attack effectiveness, eventually approaching the random-noise baseline. These results demonstrate the difficulty of circumventing our defense under the evaluated adaptive attacks.
\section{Related Work}

\paragraph{Adversarial Attacks and Defenses on VLA models.}
Recent studies have demonstrated the vulnerability of VLA models to adversarial attacks. RobotAttack~\cite{wang2025exploring} introduces adversarial patches that manipulate predicted actions and disrupt robotic trajectories. Moving beyond action-space attacks, EDPA~\cite{xu2025model} and UPA-RFAS~\cite{lu2026robots} disrupt visual representations and cross-modal alignment to achieve effective and transferable attacks. VLA-Hijack~\cite{fu2026vla} further exploits the visual self-representation of robots to improve transferability. Beyond 2D patches, Tex3D~\cite{chen2026tex3d} optimizes adversarial textures on 3D object surfaces to disrupt a robot across varying viewpoints, Universal Adversarial Object~\cite{yang2026exploiting} introduces a textured physical object to disrupt robotic control, and Phantom Menace~\cite{lu2026phantom} investigates physical sensor attacks, including light projection and laser interference. 

Compared with adversarial attacks, defenses for VLA models remain relatively underexplored. SARF~\cite{zhang2026structure} improves robustness by fine-tuning the visual encoder to rectify adversarially disrupted attention. However, its reliance on robust fine-tuning with adversarial inputs and model-specific attention mechanisms may limit its adaptability to unseen attacks and different VLA architectures. CertVLA~\cite{lu2026certvla} provides certified robustness through repeated masking and action consistency verification, but incurs substantial computational overhead due to multiple VLA inference passes.  In contrast, our method is based on visual-proprio consistency, which provides a lightweight model-agnostic defense framework. This enables our framework to generalize well without additional training.

\vspace{-0.3cm}
\paragraph{Proprioception in VLA Models.} Recent studies have explored proprioception as a source of physical priors to enhance visual understanding in VLA models. One line of research focuses on grounding visual representations in robot states. ROSA~\cite{wen2025rosa} leverages robot state estimation to improve vision-action alignment and data efficiency, while GeoProp~\cite{zhao2026geoprop} and GeoAlign~\cite{chen2026geoalign} incorporate proprioceptive information to enhance spatial grounding. Another line of research focuses on improving inference efficiency by leveraging proprioception to guide visual information processing. ThinkProprio~\cite{wang2026think} uses robot states to select action-relevant visual tokens, whereas SQAP-VLA~\cite{fang2025sqap} preserves robot-related tokens during model compression. Meanwhile, ReViP~\cite{li2026revip} and GAP~\cite{lu2026when} investigate modality imbalance between vision and proprioception to improve policy learning and generalization. These studies primarily leverage proprioception to improve policy performance or computational efficiency, while our work is the first one to explore its potential for adversarial defense.

\vspace{-0.1cm}
\section{Conclusion and Limitation}
\vspace{-0.1cm}
In this paper, we investigate the mechanisms of transferable adversarial attacks for VLAs. Based on this analysis, we build an attack-agnostic defense framework that leverages proprioception as a trusted reference to detect inconsistencies in visual robot localization. 

Our framework primarily targets transferable adversarial patches that disrupt robot-related visual representations and cannot defend against all possible attacks. For example, task-specific attacks may manipulate object perception without affecting the robot's visual representation. Defending against such attacks remains challenging even in conventional computer vision~\cite{zheng2025revisiting, nguyen2025survey}. However, task-specific attacks may require detailed knowledge of target tasks and objects, potentially limiting their transferability across different tasks. We hence focus on transferable attacks as an initial step toward more robust VLA models.
{
    \small
    \bibliographystyle{ieeenat_fullname}
    \bibliography{main}
}


\clearpage
\setcounter{page}{1}
\maketitlesupplementary

\appendix

\section{Proprioception-Guided Mask Generation}

\label{sec:appendix_mask}

We use the 3D shapes and dimensions of the robot’s components typically provided by the robot manufacturer. We select the components corresponding to the wrist, palm, and gripper, and uniformly sample surface points from them,
\begin{equation}
    \mathcal{P} = \{\mathbf{p}_{n} \in \mathbb{R}^3 \}_{n=1}^{N},
\end{equation}
where $N$ is the number of sampled surface points. Each point is represented in the local coordinate frame $L$. These points are sampled once in the local coordinate frame and provide a compact representation of the robot surface. As shown in Figure~\ref{fig:mask_generation} (left) at Section~\ref{sec:mask}.

At timestep \(t\), given the proprioceptive state $\mathbf{q}_t$, using the robot forward kinematics, we could obtain the current transformation ${}^{B} \mathbf{T}_{L,t}$ of the sampled points, where $B$ denotes the robot base frame. Each sampled point is then transformed into the camera coordinate frame $C$ as
\begin{equation}
    \bar{\mathbf{p}}^{C}_{n,t} = {}^{C}\mathbf{T}_{B}{}^{B} \mathbf{T}_{L,t}\bar{\mathbf{p}}_{n},
\end{equation}
where ${}^{C}\mathbf{T}_{B}$ denotes the calibrated camera extrinsic transformation from the robot base frame to the camera frame, and $\bar{\mathbf{p}_n}=\left[\mathbf{p}_n^{\mathrm{T}}, 1\right]$ denotes the homogeneous representation of a 3D point.

We then project each transformed 3D point onto the image plane using the camera intrinsic matrix $\mathbf{K}$,
\begin{equation}
    \mathbf{u}_{n,t} = \Pi(\mathbf{K}, \mathbf{p}^C_{n,t}),
\end{equation}
where $\Pi(\cdot)$ denotes the standard perspective projection and $\mathbf{u}_{n,t} = (u_{n,t}, v_{n,t})^\mathrm{T}$ is the corresponding image coordinate. The projected points form
\begin{equation}
    \mathcal{Q}_t = \{\mathbf{u}_{n,t}\}_{n},
\end{equation}
which represents the expected location in the camera image of the robot components, as shown in Figure~\ref{fig:mask_generation} (middle).

Based on the projected point set $\mathcal{Q}_t$, we construct a binary mask over the ViT grid cells:
\begin{equation}
m_{t,k} = \mathbb{I}\left[\mathcal{Q}_t \cap \mathcal{G}_k \neq \emptyset\right],
\quad k=1,\ldots,K,
\end{equation}
where $\mathcal{G}_k$ denotes the image region corresponding to the $k$-th grid cell, $K$ is the total number of grid cells, and $\mathbb{I}[\cdot]$ is the indicator function. Specifically, $m_{t,k}=1$ if at least one projected point falls within the $k$-th grid cell, and $m_{t,k}=0$ otherwise. The complete mask generation procedure is summarized in Alg.~\ref{alg:mask}.

\begin{algorithm}[t]
\caption{Proprioception-Guided Mask Generation}
\label{alg:mask}
\begin{algorithmic}[1]

\Require Proprioceptive state $\mathbf{q}_t$; Sampled points $\mathcal{P}$ from the robot's surface;
camera intrinsic matrix $\mathbf{K}$; camera extrinsic transformation
${}^{C}\mathbf{T}_{B}$;

\State $\mathcal{Q}_t \gets \emptyset$
\Comment{Projected robot surface points}

    \State Obtain ${}^{B}\mathbf{T}_{L,t}$ from $\mathbf{q}_t$
    using the forward kinematics

    \For{each $\mathbf{p}_{n} \in \mathcal{P}$}

        \State $\bar{\mathbf{p}}_{n}
        \gets [\mathbf{p}_{n}^{\top},1]^{\top}$
        \Comment{Homogeneous coordinate}

        \State $\bar{\mathbf{p}}^{C}_{n,t}
        \gets
        {}^{C}\mathbf{T}_{B}
        {}^{B}\mathbf{T}_{L,t}
        \bar{\mathbf{p}}_{n}$
        \Comment{Transform the point into the camera frame}

        \State $\mathbf{p}^{C}_{n,t}
        \gets
        \bar{\mathbf{p}}^{C}_{n,t}[1\!:\!3]$
        \Comment{$\mathbf{p}^{C}_{n,t}=(X,Y,Z)^{\top}$}

        \State $\mathbf{u}_{n,t}
        \gets
        \Pi(\mathbf{K},\mathbf{p}^{C}_{n,t})$
        \Comment{Perspective projection}

        \State $\mathcal{Q}_t
        \gets
        \mathcal{Q}_t \cup
        \{\mathbf{u}_{n,t}\}$

    \EndFor

    \State Initialize $\mathbf{m}_t \gets \mathbf{0} \in \{0,1\}^K$
    \For{each projected point $\mathbf{u} \in \mathcal{Q}_t$}
        \State $k \gets \operatorname{GridIndex}(\mathbf{u})$
        \Comment{Find the grid containing $\mathbf{u}$}
        \State $m_{t,k} \gets 1$
    \EndFor

\State \Return $\mathbf{m}_t
=
[m_{t,1},\ldots,m_{t,K}]$

\end{algorithmic}
\end{algorithm}

\section{Defense Framework Summarization}

We summarize our defense pipeline in Alg.~\ref{alg:defense}.

\begin{algorithm}[t]
\caption{Visual-Proprioceptive Consistency Defense}
\label{alg:defense}
\begin{algorithmic}[1]
\Require Observation $(I_t,\ell_t,\mathbf{q}_t)$,
VLA policy $\pi_\theta$, detection threshold $\delta$,
maximum masking budget $B$

\Statex \textbf{Detection}
\State $\mathbf{m}_t \gets
    \textsc{GenerateMask}(\mathbf{q}_t)$
    \Comment{Alg.~\ref{alg:mask}}
\State $\mathbf{Z}_t \gets f_{\mathrm{vis}}(I_t)$
    \Comment{Extract visual features}
\State $d_t \gets
    D_{\mathrm{PV}}(\mathbf{Z}_t,\mathbf{m}_t)$

\If{$d_t \leq \delta$}
    \State \Return $\pi_\theta(I_t,\ell_t,\mathbf{q}_t)$
    \Comment{No attack detected}
\EndIf

\Statex \textbf{Patch Localization}
\State $\mathbf{s}_t \gets
    \textsc{SmoothGrad}(D_{\mathrm{PV}}, I_t)$
\Comment{Pixel-level attribution}
\State $(k_1,\ldots,k_R) \gets
    \textsc{RankTokens}(\mathbf{s}_t,\mathbf{m}_t)$
\Comment{Aggregate attribution per grid, exclude tokens with $m_{t,k}=1$, and rank the remaining tokens in descending order}

\Statex \textbf{Patch Mitigation}
\State $\mathcal{H}_t \gets \emptyset$

\For{$r=1,\ldots,B$}
    \State $\mathcal{H}_t \gets
        \mathcal{H}_t \cup \{k_r\}$

    \State $\widetilde{\mathbf{Z}}_t \gets
        f_{\mathrm{vis}}(I_t;\mathcal{H}_t)$
    \Comment{Attention masking $\mathcal{H}_t$}

    \State $\widetilde{d}_t \gets
        D_{\mathrm{PV}}(\widetilde{\mathbf{Z}}_t,\mathbf{m}_t)$

    \If{$\widetilde{d}_t \leq \delta$}
        \State \textbf{break}
    \EndIf
\EndFor

\State $\mathbf{A}_t \gets
    \pi_\theta^{\mathcal{H}_t}
    (I_t,\ell_t,\mathbf{q}_t)$
\State \Return $\mathbf{A}_t$

\end{algorithmic}
\end{algorithm}

\section{Implementation details}

We sample 1,000 frames for each LIBERO task suite from the training demonstrations, balanced across the ten tasks. We set the temperature parameter be $0.1\kappa$, where $\kappa$ is the mean of the robot-likeness score $\mathcal{R}_{t,k}$ across the training data. And for the SmoothGrad-based attribution, we set $\sigma=0.15$ and 16 noise samples. And for the robot feature preserving attack, the objective is
$$
    \mathcal{L}_{\text{robot}} = \mathcal{L}_{\mathrm{UADA}} + \lambda\sum_{\mathbf{I}_t\in\mathcal{D}}\frac{1}{|\mathcal{D}||M_t|}\sum_{k\in M_t} \frac{\lVert \mathbf{z}_{t,k}' - \mathbf{z}_{t,k}\rVert_2^2}{\lVert \mathbf{z}_{t,k}\rVert_2^2},
$$
where $M_t$ is the mask that indicates the indices of robot-related visual tokens. And for the detection-aware attack, the objective is
$$
    \mathcal{L}_{\text{detect}} = \mathcal{L}_{\mathrm{UADA}} + \lambda D_{PV},
$$
where $D_{PV}$ is our proposed visual-proprio consistency score for attack detection.

\section{Supplementary experiment results}

\subsection{Threshold Calibration}

For practical deployment, a detection threshold is required to determine whether an observation is under attack. To maintain the attack-agnostic nature of our method, we select this threshold without using any adversarial attack data. Moreover, to improve system robustness, we aim to minimize false alarms caused by benign visual perturbations. Specifically, we augment clean training images with randomly generated noise patches and select a threshold that keeps the false-positive rate on these non-adversarial patches below 5\%. Using this fixed threshold, we report the true-positive rate (TPR) on adversarial observations and the false-positive rate (FPR) on clean observations. As shown in Table~\ref{tab:threshold}, our attack-agnostic thresholds achieve average TPRs of 97.8\% and 93.8\% for OpenVLA and $\pi_0$, respectively, while maintaining average FPRs of only 1.7\% and 0.9\%, with no FPR exceeding 3.0\% across all four suites.

\begin{table}[htbp]
  \caption{TPR and FPR with selected threshold.}
  \vspace{-0.3cm}
  \label{tab:threshold}
  \centering
  \begin{tabular}{@{}ccccc@{}}
    \toprule
    & \multicolumn{2}{c}{OpenVLA}
    & \multicolumn{2}{c}{$\pi_0$} \\
    \cmidrule(lr){2-3}
    \cmidrule(l){4-5}
    Suite & TPR & FPR & TPR & FPR \\
    \midrule
    Spatial & 0.998 & 0.028 & 0.958 & 0.000 \\
    Goal    & 0.995 & 0.015 & 0.917 & 0.007 \\
    Object  & 1.000 & 0.000 & 0.959 & 0.000 \\
    Long    & 0.918 & 0.023 & 0.916 & 0.030 \\
    \bottomrule
  \end{tabular}
\end{table}

\subsection{Patch Placement Experiment}

Since the proposed detection score $D_{\mathrm{PV}}$ measures how far the robot-like tokens lie from the trusted support. Whether a patch is detected may depend on where it is placed relative to the robot. To measure this dependence, we move each patch over 50 different positions. For each placement we record the detector's score, and we report the detection capability of the patch's distance to the robot. The results are shown in Table~\ref{tab:placement}, our detection method remains effective even when the patch is placed close to the robot.

\begin{table}[t]
\centering
\small
\setlength{\tabcolsep}{4pt}
\caption{Detection AUROC by the patch's distance to the robot region in tokens.}
\label{tab:placement}
\begin{tabular}{llcccc}
\toprule
Suite & Attack & $[1,3)$ & $[3,5)$ & $[5,7)$ & $\ge7$ \\
\midrule
\multicolumn{6}{l}{\textbf{OpenVLA}} \\
\midrule
Spatial & VLA-Hijack  & 0.999 & 1.000 & 1.000 & 1.000 \\
 & UPA-RFAS & 1.000 & 1.000 & 1.000 & 1.000 \\
 & EDPA & 1.000 & 1.000 & 1.000 & 1.000 \\
 & UADA & 1.000 & 1.000 & 1.000 & 1.000 \\
 & TMA &  0.999 & 1.000 & 1.000 & 1.000 \\
 & UMA &  1.000 & 1.000 & 1.000 & 1.000 \\
 & UPA & 0.993 & 0.999 & 1.000 & 1.000 \\
\cmidrule{1-6}
Object & VLA-Hijack & 1.000 & 1.000 & 1.000 & 1.000 \\
 & UPA-RFAS & 1.000 & 1.000 & 1.000 & 1.000 \\
 & EDPA & 1.000 & 1.000 & 1.000 & 1.000 \\
 & UADA & 1.000 & 1.000 & 1.000 & 1.000 \\
 & TMA & 1.000 & 1.000 & 1.000 & 1.000 \\
 & UMA & 1.000 & 1.000 & 1.000 & 1.000 \\
 & UPA & 1.000 & 1.000 & 1.000 & 1.000 \\
\cmidrule{1-6}
Goal & VLA-Hijack & 0.998 & 0.999 & 1.000 & 1.000 \\
 & UPA-RFAS & 1.000 & 1.000 & 1.000 & 1.000 \\
 & EDPA & 1.000 & 1.000 & 1.000 & 1.000 \\
 & UADA & 0.999 & 1.000 & 1.000 & 1.000 \\
 & TMA & 0.999 & 1.000 & 1.000 & 1.000 \\
 & UMA & 0.999 & 1.000 & 1.000 & 1.000 \\
 & UPA & 0.994 & 0.999 & 1.000 & 1.000 \\
\cmidrule{1-6}
LIBERO-10 & VLA-Hijack & 0.923 & 0.936 & 0.946 & 0.969 \\
 & UPA-RFAS & 0.981 & 0.985 & 0.988 & 0.991 \\
 & EDPA & 0.970 & 0.980 & 0.985 & 0.989 \\
 & UADA & 0.982 & 0.986 & 0.988 & 0.992 \\
 & TMA & 0.974 & 0.983 & 0.986 & 0.989 \\
 & UMA & 0.980 & 0.985 & 0.987 & 0.991 \\
 & UPA & 0.956 & 0.977 & 0.983 & 0.987 \\
\midrule
\multicolumn{6}{l}{\textbf{$\pi_0$}} \\
\midrule
Spatial & VLA-Hijack & 0.946 & 0.960 & 0.981 & 0.984 \\
 & UPA-RFAS & 0.985 & 0.996 & 0.991 & 0.990 \\
 & EDPA  & 0.980 & 0.989 & 0.985 & 0.987 \\
\cmidrule{1-6}
Object & VLA-Hijack & 0.939 & 0.947 & 0.988 & 0.989 \\
 & UPA-RFAS & 0.981 & 0.999 & 0.999 & 0.997 \\
 & EDPA & 0.979 & 0.999 & 1.000 & 0.999 \\
\cmidrule{1-6}
Goal & VLA-Hijack & 0.937 & 0.955 & 0.957 & 0.989 \\
 & UPA-RFAS & 0.975 & 0.984 & 0.984 & 0.997 \\
 & EDPA & 0.982 & 0.988 & 0.986 & 0.997 \\
\cmidrule{1-6}
LIBERO-10 & VLA-Hijack & 0.902 & 0.917 & 0.937 & 0.952 \\
 & UPA-RFAS & 0.918 & 0.927 & 0.934 & 0.967 \\
 & EDPA & 0.935 & 0.922 & 0.939 & 0.972 \\
\bottomrule
\end{tabular}
\end{table}

\subsection{Visualization of Patch Localization}

\begin{figure*}
    \centering
    \includegraphics[width=0.85\linewidth]{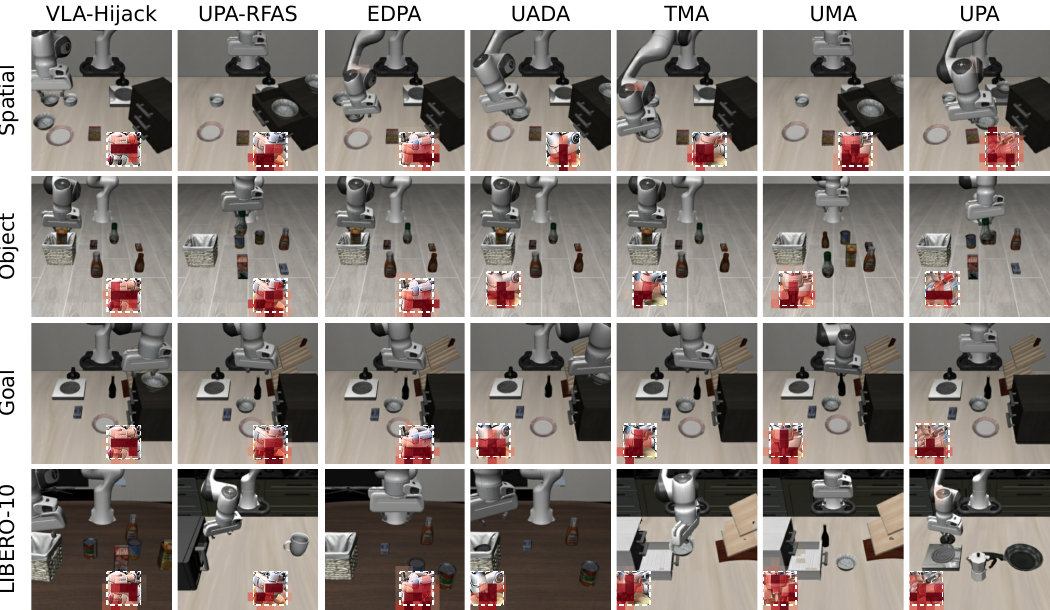}
    \caption{Patch localization for OpenVLA.}
    \label{fig:localization_openvla}
\end{figure*}

\begin{figure}
    \centering
    \includegraphics[width=0.85\linewidth]{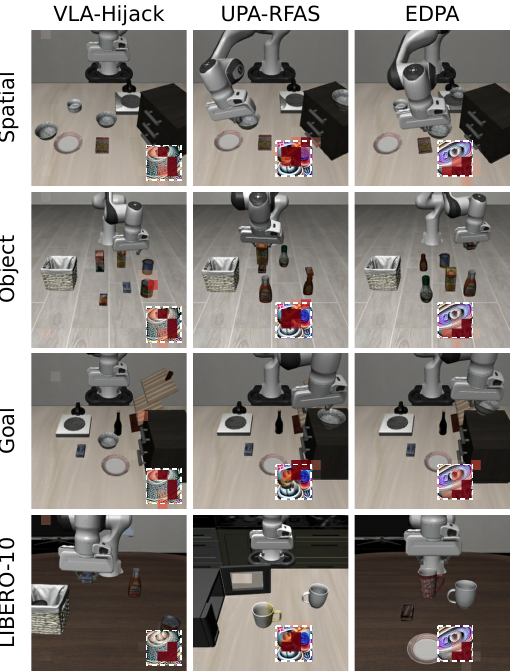}
    \caption{Patch localization for $\pi_0$.}
    \label{fig:localization_pi0}
\end{figure}

Figure~\ref{fig:localization_openvla} and Figure~\ref{fig:localization_pi0} show the patch localization of the SmoothGrad-based attribution method. We show the attribution scores of each visual token for OpenVLA and $\pi_0$ across different patch attacks. Results show that the attribution method can accurately localize the patches based on our visual-proprioceptive consistency score.

\subsection{Camera Shift Detection}

\begin{figure}
    \centering
    \includegraphics[width=\linewidth]{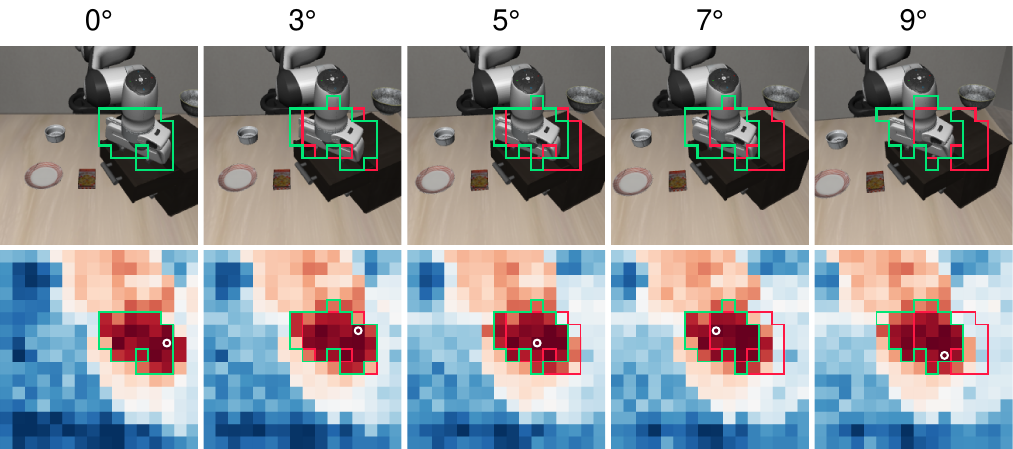}
    \caption{Camera calibration shift detection. \textbf{Red box:} Masks rendered from staled camera parameters. \textbf{Green box:} Masks rendered from correct camera parameters.}
    \label{fig:viewpoint_shift}
\end{figure}

Beyond adversarial defense, our visual-proprio consistency principle can be extended to detect other anomalies in robotic systems, such as unexpected camera displacement, temporal misalignment caused by camera latency. For VLA deployment, it is very common for the camera viewpoint to be changed due to accidents such as collisions, and this anomaly may affect the performance of VLA models. 
This camera shifts cause input image shifts while the calibration used for mask generation remains unchanged. Therefore, there will be an inconsistency between the visual location and proprioceptive state, which can be detected by our method. In order to test the camera calibration detection capability, we leverage LIBERO-Plus benchmark, and synthesis viewpoint rotations from LIBERO-Spatial suite. As shown in Figure~\ref{fig:viewpoint_shift}, we could see that after camera viewpoint shift, there is a gap between the visual location and mask rendered from staled camera calibration. Table~\ref{tab:view_shift} further shows the detection capability under different degrees of camera viewpoint change.

\begin{table}[htbp]
  \caption{Detection capability (AUROC) under different degrees of camera viewpoint change.}
  \vspace{-0.3cm}
  \label{tab:view_shift}
  \centering
  \begin{tabular}{@{}ccccc@{}}
    \toprule
    View Shift & $3^{\circ}$ & $5^{\circ}$ & $7^{\circ}$ & $9^{\circ}$ \\
    \midrule
    OpenVLA & 0.849 & 0.948 & 0.981 & 0.954\\
    $\pi_0$ &0.875 & 0.949 & 0.954 & 0.937\\
    \bottomrule
    \end{tabular}
\end{table}

\end{document}